\documentclass[11pt,a4paper]{article}
\usepackage[hyperref]{acl2021}
\usepackage{times}
\usepackage{latexsym}
\usepackage{amssymb}
\usepackage{bm}
\usepackage{amsmath}

\usepackage{booktabs}
\usepackage{graphics}
\usepackage{multirow}
\usepackage{color}
\usepackage{xcolor}
\usepackage{graphicx}

\usepackage{microtype}

\aclfinalcopy 
\def\aclpaperid{3981} 

\title{CLINE: Contrastive Learning with Semantic Negative Examples for Natural Language Understanding}

\author{
Dong Wang$^{1,2}\thanks{\quad Equal contribution. This work was mainly done when Dong Wang was an intern at Tencent AI Lab.}$\hspace{0.5em}, 
Ning Ding$^{1,2*}$, 
Piji Li$^{3}$\thanks{\quad Corresponding authors.}\hspace{0.5em},
Hai-Tao Zheng$^{1,2}$\footnotemark[2]\\
$^{1}$Department of Computer Science and Technology, Tsinghua University\\
$^{2}$Tsinghua ShenZhen International Graduate School, Tsinghua University\\
$^{3}$Tencent AI Lab\\
\tt{\{wangd18,dingn18\}@mails.tsinghua.edu.cn}\\
\tt{pijili@tencent.com,zheng.haitao@sz.tsinghua.edu.cn}
}

\date{}

\begin{document}
\maketitle
\begin{abstract}
Despite pre-trained language models have proven useful for learning high-quality semantic representations, these models are still vulnerable to simple perturbations. Recent works aimed to improve the robustness of pre-trained models mainly focus on adversarial training from perturbed examples with similar semantics, neglecting the utilization of different or even opposite semantics. Different from the image processing field, the text is discrete and few word substitutions can cause significant semantic changes. To study the impact of semantics caused by small perturbations, we conduct a series of pilot experiments and surprisingly find that adversarial training is useless or even harmful for the model to detect these semantic changes. To address this problem, we propose Contrastive Learning with semantIc Negative Examples (CLINE), which constructs semantic negative examples unsupervised to improve the robustness under semantically adversarial attacking. By comparing with similar and opposite semantic examples, the model can effectively perceive the semantic changes caused by small perturbations. Empirical results show that our approach yields substantial improvements on a range of sentiment analysis, reasoning, and reading comprehension tasks. And CLINE also ensures the compactness within the same semantics and separability across different semantics in sentence-level.


\end{abstract}

\section{Introduction}
\label{sec:intro}

\begin{table}[t!]
\begin{center}
\scalebox{0.9}{
\begin{tabular}{p{0.45\columnwidth} cc}
\toprule
\textbf{Sentence} & \textbf{Label} & \textbf{Predict} \\
\midrule
creepy but ultimately unsatisfying thriller & \textcolor{blue}{Negative} & \textcolor{blue}{Negative} \\
\midrule
creepy but \underline{lastly} unsatisfying thriller & \textcolor{blue}{Negative} & \textcolor{red}{Positive} \\
\midrule
creepy but ultimately \underline{satisfying} thriller & \textcolor{red}{Positive} & \textcolor{blue}{Negative} \\
\bottomrule
\end{tabular}}
\end{center}
\caption{\label{tab:examples} An adversarial example of sentiment analysis in movie reviews. And the prediction results are from the BERT (base version with 12 layers). }
\end{table}

Pre-trained language models (PLMs) such as BERT \cite{DevlinCLT19} and RoBERTa \cite{liu2019roberta} have been proved to be an effective way to improve various natural language processing tasks. However, recent works show that PLMs suffer from poor robustness when encountering adversarial examples \cite{JinJZS20,LiMGXQ20,GargR20,ZangQYLZLS20,Lin0YOCGM20}. As shown in Table \ref{tab:examples}, the BERT model can be fooled easily just by replacing \emph{ultimately} with a similar word \emph{lastly}.

To improve the robustness of PLMs, recent studies attempt to adopt adversarial training on PLMs, which applies gradient-based perturbations to the word embeddings during training \cite{MiyatoDG17,ZhuCGSGL20,JiangHCLGZ20} or adds high-quality adversarial textual examples to the training phase \cite{WangB18,MichelLNP19}.
The primary goal of these adversarial methods is to keep the label unchanged when the input has small changes. 
These models yield promising performance by constructing high-quality perturbated examples and adopting adversarial mechanisms. 
However, due to the discrete nature of natural language, in many cases, small perturbations can cause significant changes in the semantics of sentences.
As shown in Table \ref{tab:examples}, negative sentiment can be turned into a positive one by changing only one word, but the model can not recognize the change.
Some recent works create contrastive sets \cite{KaushikHL20,gardner2020evaluating}, which manually perturb the test instances in small but meaningful ways that change the gold label. In this paper, we denote the perturbated examples without changed semantics as adversarial examples and the ones with changed semantics as contrastive examples, and most of the methods to improve robustness of PLMs mainly focus on the former examples, little study pays attention to the semantic negative examples.


The phenomenon makes us wonder \emph{can we train a BERT that is both defensive against adversarial attacks and sensitive to semantic changes by using both adversarial and contrastive examples?} To answer that, we need to assess if the current robust models are meanwhile semantically sensitive. We conduct sets of pilot experiments (Section~\ref{sec:analysis}) to compare the performances of vanilla PLMs and adversarially trained PLMs on the contrastive examples. We observe that while improving the robustness of PLMs against adversarial attacks, the performance on contrastive examples drops.




To train a robust semantic-aware PLM, we propose Contrastive Learning with semantIc Negative Examples (CLINE).
CLINE is a simple and effective method to generate adversarial and contrastive examples and contrastively learn from both of them.
The contrastive manner has shown effectiveness in learning sentence representations~\cite{luo2020capt,wu2020clear,Gao2021simcse}, yet these studies neglect the generation of negative instances. In CLINE, we use external semantic knowledge, i.e., WordNet~\cite{Miller95}, to generate adversarial and contrastive examples by unsupervised replacing few specific representative tokens.  
Equipped by replaced token detection and contrastive objectives, our method gathers similar sentences with semblable semantics and disperse ones with different even opposite semantics, simultaneously improving the robustness and semantic sensitivity of PLMs. We conduct extensive experiments on several widely used text classification benchmarks to verify the effectiveness of CLINE. To be more specific, our model achieves +1.6\% absolute improvement on 4 contrastive test sets and +0.5\% absolute improvement on 4 adversarial test sets compared to RoBERTa model \cite{liu2019roberta}. That is, with the training on the proposed objectives, CLINE simultaneously gains the robustness of adversarial attacks and sensitivity of semantic changes\footnote{The source code of CLINE will be publicly available at \url{https://github.com/kandorm/CLINE}}.



\section{Pilot Experiment and Analysis}
\label{sec:analysis}
To study how the adversarial training methods perform on the adversarial set and contrastive set, we first conduct pilot experiments and detailed analyses in this section.

\subsection{Model and Datasets}
\label{ssec:pdm}

There are a considerable number of studies constructing adversarial examples to attack large-scale pre-trained language models, of which we select a popular method, TextFooler~\cite{JinJZS20}, as the word-level adversarial attack model to construct adversarial examples.
Recently, many researchers create contrastive sets to more accurately evaluate a model’s true linguistic capabilities \cite{KaushikHL20,gardner2020evaluating}. 
Based on these methods, the following datasets are selected to construct adversarial and contrastive examples in our pilot experiments and analyses:

{\bf IMDB} \cite{MaasDPHNP11} is a sentiment analysis dataset and the task is to predict the sentiment (positive or negative) of a movie review.

{\bf SNLI} \cite{BowmanAPM15} is a natural language inference dataset to judge the relationship between two sentences: whether the second sentence can be derived from entailment, contradiction, or neutral relationship with the first sentence.

To improve the generalization and robustness of language models, many adversarial training methods that minimize the maximal risk for label-preserving input perturbations have been proposed, and we select an adversarial training method FreeLB \cite{ZhuCGSGL20} for our pilot experiment. We evaluate the vanilla BERT \cite{DevlinCLT19} and RoBERTa \cite{liu2019roberta}, and the FreeLB version on the adversarial set and contrastive set.

\begin{table*}
\centering
\scalebox{0.95}{
\begin{tabular}{l|c|cc|cc}
\toprule
\multirow{2}{*}{Model} & \multirow{2}{*}{Method} & \multicolumn{2}{c|}{IMDB} & \multicolumn{2}{c}{SNLI} \\ 
& & Adv & Rev & Adv & Rev \\ \midrule

\multirow{2}{*}{BERT-base} & Vanilla & 88.7 & 89.8 & 48.6 & 73.0  \\
& FreeLB & 91.9 ($\color{teal}{+ 3.2}$) & 87.7 ($\color{red}{- 2.1}$) & 56.1 ($\color{teal}{+ 7.5}$) & 71.4 ($\color{red}{- 1.6}$) \\
\midrule
\multirow{2}{*}{RoBERTa-base} & Vanilla & 93.9 &  93.0 & 55.1 &  75.2 \\
& FreeLB & { 95.2 ($\color{teal}{+ 1.3}$)} & 92.6 ($\color{red}{- 0.4}$) &  58.1 ($\color{teal}{+ 3.0}$) & 74.6 ($\color{red}{- 0.6}$)  \\
\bottomrule
\end{tabular}}
\caption{\label{tab:pexp} Accuracy (\%) on the adversarial set (Adv) compared to the contrastive set (Rev) of Vanilla models and adversarially trained models. }
\end{table*}

\subsection{Result Analysis}
\label{ssec:pra}
Table \ref{tab:pexp} shows a detailed comparison of different models on the adversarial test set and the contrast test set. From the results, we can observe that, compared to the vanilla version, the adversarial training method FreeLB achieves higher accuracy on the adversarial sets, but suffers a considerable performance drop on the contrastive sets, especially for the BERT. The results are consistent with the intuition in Section~\ref{sec:intro}, and also demonstrates that adversarial training is not suitable for the contrastive set and even brings negative effects. 
Intuitively, adversarial training tends to keep labels unchanged while the contrastive set tends to make small but label-changing modifications. The adversarial training and contrastive examples seem to constitute a natural contradiction, revealing that additional strategies need to be applied to the training phase for the detection of the fine-grained changes of semantics.
We provide a case study in Section \ref{ssec:pcase}, which further shows this difference.

\begin{table}
\centering
\scalebox{0.9}{
\begin{tabular}{p{0.9\columnwidth}}
\toprule
\textbf{IMDB Contrastive Set} \\
\midrule
Jim Henson's Muppets were \textcolor{red}{a favorite of mine since childhood}. This film on the other hand \textcolor{blue}{makes me feel dizziness} in my head. You will see \textcolor{red}{cameos by the then New York City Mayor Ed Koch}. Anyway, the film turns 25 this year and I hope the kids of today will \textcolor{red}{learn to appreciate the lightheartedness} of the early Muppets Gang \textcolor{blue}{over this}. It might \textcolor{red}{be worth watching for kids} but definitely \textcolor{blue}{not} for knowledgeable adults like myself.

Label: \textcolor{blue}{Negative}

Prediction: \textcolor{red}{Positive} \\
\bottomrule
\end{tabular}}
\caption{\label{tab:pcase} Wrong predictions made by the FreeLB version of BERT on the contrastive set. }
\end{table}

\subsection{Case Study}
\label{ssec:pcase}
To further understand why the adversarial training method fails on the contrastive sets, we carry out a thorough case study on IMDB. 
The examples we choose here are predicted correctly by the vanilla version of BERT but incorrectly by the FreeLB version. For the example in Tabel \ref{tab:pcase}, we can observe that many parts are expressing positive sentiments (red part) in the sentence, and a few parts are expressing negative sentiments (blue parts). Overall, this case expresses negative sentiments, and the vanilla BERT can accurately capture the negative sentiment of the whole document.
However, the FreeLB version of BERT may take the features of negative sentiment as noise and predict the whole document as a positive sentiment.  This result indicates that the adversarially trained BERT could be fooled in a reversed way of traditional adversarial training.
From this case study, we can observe that the adversarial training methods may not be suitable for these semantic changed adversarial examples, and to the best of our knowledge, there is no defense method for this kind of adversarial attack. Thus, it is crucial to explore the appropriate methods to learn changed semantics from semantic negative examples.

\section{Method}
\label{sec:method}

\begin{figure*}[t]
    \centering
    \includegraphics[width = 0.98\linewidth]{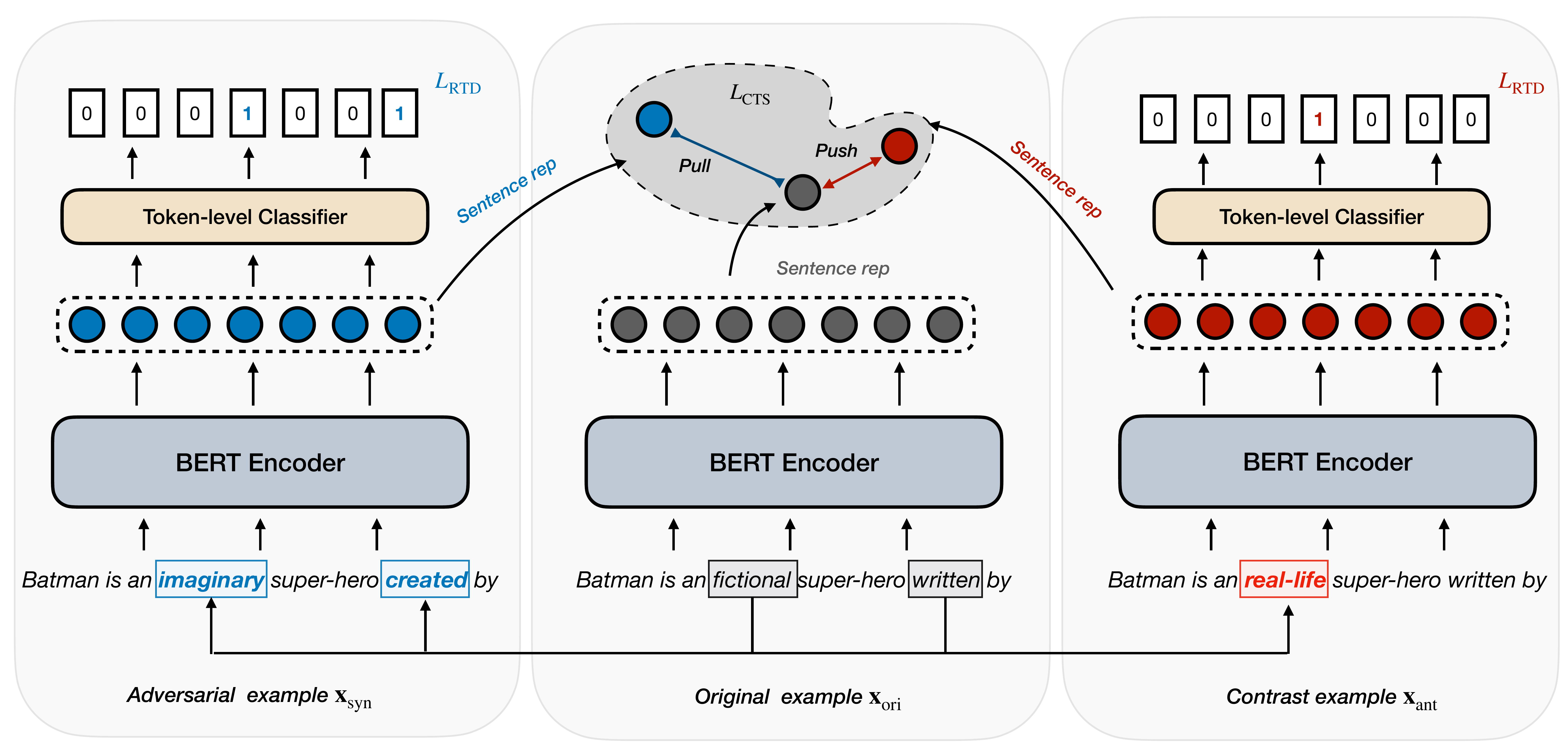}
    \caption{An illustration of our model, note that we use the embedding of \texttt{[CLS]} as the sentence representation.}
    \label{fig:model}
\end{figure*}

As stated in the observations in Section~\ref{sec:analysis}, we explore strategies that could improve the sensitivity of PLMs. In this section, we present CLINE, a simple and effective method to generate the adversarial and contrastive examples and learn from both of them. We start with the generation of adversarial and contrastive examples in Section~\ref{sec:gen}, and then introduce the learning objectives of CLINE in Section~\ref{sec:obj}.

\subsection{Generation of Examples}
\label{sec:gen}

We expect that by contrasting sentences with the same and different semantics, our model can be more sensitive to the semantic changes. To do so, we adopt the idea of contrastive learning, which aims to learn the representation by concentrating positive pairs and pushing negative pairs apart. Therefore it is essential to define appropriate positive and negative pairs. In this paper, we regard sentences with the same semantics as positive pairs and sentences with opposite semantics as negative pairs. Some works \cite{AlzantotSEHSC18,TanJKS20,wu2020clear} attempt to utilize data augmentation (such as synonym replacement, back translation, etc) to generate positive instances, but few works pay attention to the negative instances. And it is difficult to obtain opposite semantic instances for textual examples.

Intuitively, when we replace the representative words in a sentence with its antonym, the semantic of the sentence is easy to be irrelevant or even opposite to the original sentence. As shown in Figure \ref{fig:model}, given the sentence ``Batman is an fictional super-hero written by'', we can replace ``fictional'' with its antonym ``real-life'', and then we get a counterfactual sentence ``Batman is an real-life super-hero written by''. The latter contradicts the former and forms a negative pair with it. 

We generate two sentences from the original input sequence $\bm{x}^{\rm ori}$, which express substantially different semantics but have few different words.
One of the sentences is semantically close to $\bm{x}^{\rm ori}$ (denoted as $\bm{x}^{\rm syn}$), while the other is far from or even opposite to $\bm{x}^{\rm ori}$ (denoted as $\bm{x}^{\rm ant}$).
In specific, we utilize spaCy\footnote{\url{https://github.com/explosion/spaCy}} to conduct segmentation and POS for the original sentences, extracting verbs, nouns, adjectives, and adverbs. $\bm{x}^{\rm syn}$ is generated by replacing the extracted words with synonyms, hypernyms and morphological changes, and $\bm{x}^{\rm ant}$ is generated by replacing them with antonyms and random words.  For $\bm{x}^{\rm syn}$, about 40\% tokens are replaced. For $\bm{x}^{\rm ant}$, about 20\% tokens are replaced.

\subsection{Training Objectives}
\label{sec:obj}

CLINE trains a neural text encoder (i.e., deep Transformer) $\texttt{E}_\phi$ parameterized by $\phi$ that maps a sequence of input tokens $\bm{x}=[x_1,...,x_T]$ to a sequence of representations $\bm{h}=[h_1,.., h_T], h_{i \in [1:T]} \in \mathbb{R}^d$, where $d$ is the dimension:
\begin{equation}
	\bm{h} = \texttt{E}_\phi(\bm{x}).
\end{equation}

\noindent
\textbf{Masked Language Modeling Objective} With random tokens masked by special symbols $\texttt{[MASK]}$, the input sequence is partially corrupted. Following BERT~\cite{DevlinCLT19}, we adopt the masked language model objective (denoted as $\mathcal{L}_{\text{MLM}}$), which reconstructs the sequence by predicting the masked tokens.


\noindent
\textbf{Replaced Token Detection Objective} On the basis of $\bm{x}^{\rm syn}$ and $\bm{x}^{\rm ant}$, we adopt an additional classifier $\mathcal{C}$ for the two generated sequences and detect which tokens are replaced by conducting two-way classification with a sigmoid output layer:
\begin{equation}
    p(\bm{x}^{\rm syn}, t) = \text{sigmoid}(w^\top h^{\rm syn}_t),
\end{equation}
\begin{equation}
    p(\bm{x}^{\rm ant}, t) = \text{sigmoid}(w^\top h^{\rm ant}_t).
\end{equation}
The loss, denoted as $\mathcal{L}_{\text{RTD}}$ is computed by:

\begin{equation}
\begin{split}
    \mathcal{L}_{\text{RTD}} &= \sum_{\bm{x}' \in \{x^{\rm syn}, x^{\rm ant}\}} -\sum_{t=1}^T \delta_t \text{log}\ p(\bm{x}',t) \\ &-  (1-\delta_t) \text{log}(1-  p(\bm{x}', t)),
    \end{split}
\end{equation}
where $\delta_t = 1$ when the token $x_t$ is corrupted, and $\delta_t =0$ otherwise.

\noindent
\textbf{Contrastive Objective} The intuition of CLINE is to accurately predict if the semantics are changed when the original sentences are modified. In other words, in feature space, the metric between $\bm{h}^{\rm ori}$ and $\bm{h}^{\rm syn}$ should be close and the metric between $\bm{h}^{\rm ori}$ and $\bm{h}^{\rm ant}$ should be far. Thus, we develop a contrastive objective, where ($\bm{x}^{\rm ori}$, $\bm{x}^{\rm syn}$) is considered a positive pair and ($\bm{x}^{\rm ori}$, $\bm{x}^{\rm ant}$) is negative. We use $h_c$ to denote the embedding of the special symbol \texttt{[CLS]}. In the training of CLINE, we follow the setting of RoBERTa~\cite{liu2019roberta} to omit the next sentence prediction (NSP) objective since previous works have shown that NSP objective can hurt the performance on the downstream tasks \cite{liu2019roberta,JoshiCLWZL20}. Alternatively, adopt the embedding of \texttt{[CLS]} as the sentence representation for a contrastive objective.
The metric between sentence representations is calculated as the dot product between \texttt{[CLS]} embeddings:
\begin{equation}
    f(\bm{x}^*, \bm{x}') = \text{exp}(h_c^{*\top} h'_c).
\end{equation}
Inspired by InfoNCE, we define an objective $\mathcal{L}_{\rm cts}$ in the contrastive manner:
\begin{equation}
    \mathcal{L}_{\rm cts}\! =\! -\sum_{\bm{x}\in \mathcal{X}} \text{log} \frac{f(\bm{x}^{\rm ori}, \bm{x}^{\rm syn})}{f(\bm{x}^{\rm ori}, \bm{x}^{\rm syn}) + f(\bm{x}^{\rm ori}, \bm{x}^{\rm ant})}.
\end{equation}
Note that different from some contrastive strategies that usually randomly sample multiple negative examples, we only utilize one $\bm{x}_{\rm ant}$ as the negative example for training. This is because the primary goal of our pre-training objectives is to improve the robustness under semantically adversarial attacking. And we only focus on the negative sample (i.e., $\bm{x}_{\rm ant}$) that is generated for our goal, instead of arbitrarily sampling other sentences from the pre-training corpus as negative samples.

Finally, we have the following training loss:
\begin{equation}
    \mathcal{L}\! =\! \lambda_1 \mathcal{L}_{\text{MLM}} + \lambda_2 \mathcal{L}_{\text{RTD}} + 
    \lambda_3 \mathcal{L}_{\rm cts},
\end{equation}
where $\lambda_i$ is the task weighting learned by training.




\section{Experiments}
\label{sec:exp}

We conduct extensive experiments and analyses to evaluate the effectiveness of CLINE. In this section, we firstly introduce the implementation (Section~\ref{sec:imp}) and the datasets (Section~\ref{sec:data}) we used, then we introduce the experiments on contrastive sets (Section~\ref{sec:expcon}) and adversarial sets (Section~\ref{sec:expadv}), respectively. Finally, we conduct the ablation study (Section~\ref{ssec:as}) and analysis about sentence representation (Section ~\ref{ssec:ssr}).

\subsection{Implementation}
\label{sec:imp}

To better acquire the knowledge from the existing pre-trained model, we did not train from scratch but the official RoBERTa-base model. We train for 30K steps with a batch size of 256 sequences of maximum length 512 tokens. We use Adam with a learning rate of 1e-4, $\beta_1=0.9$, $\beta_2=0.999$, $\epsilon=$1e-8, L2 weight decay of 0.01, learning rate warmup over the first 500 steps, and linear decay of the learning rate. We use 0.1 for dropout on all layers and in attention. The model is pre-trained on 32 NVIDIA Tesla V100 32GB GPUs. Our model is pre-trained on a combination of BookCorpus \cite{ZhuKZSUTF15} and English Wikipedia datasets, the data BERT used for pre-training.

\begin{table*}[t!]
\begin{center}
\scalebox{0.9}{
\begin{tabular}{l|ccc|ccc|ccc|ccc}
  \toprule
  \multirow{2}{*}{Model} & \multicolumn{3}{c}{IMDB} & \multicolumn{3}{|c}{PERSPECTRUM} & \multicolumn{3}{|c}{BoolQ} & \multicolumn{3}{|c}{SNLI} \\
  & Ori & Rev & Con & Ori & Rev & Con & Ori & Rev & Con & Ori & Rev & Con \\
  \midrule
  BERT & 92.2 & 89.8 & 82.4 & 74.7 & 72.8 & 57.6 & 60.9 & 57.6 & 36.1 & 89.8 & 73.0 & 65.1 \\
  RoBERTa & 93.6 & 93.0 & 87.1 & 80.6 & 78.8 & 65.0 & 69.6 & 60.6 & 43.9 & 90.8 & 75.2 & 67.8  \\
  \midrule
  CLINE    & \bf 94.5 & \bf 93.9 & \bf 88.5 & \bf 81.6 & \bf 80.2 & \bf 72.2 & \bf 73.9 & \bf 63.9 & \bf 47.8 & \bf 91.3 & \bf 76.0 & \bf 69.2 \\
  \bottomrule
\end{tabular}}
\end{center}
\caption{\label{tab:contrast} Accuracy on the original test set (Ori) and contrastive test set (Rev). Contrast consistency (Con) is a metric of whether a model makes correct predictions on every element in both the original test set and the contrastive test set.}
\end{table*}

\subsection{Datasets}
\label{sec:data}

We evaluate our model on six text classification tasks:
\begin{itemize}
    \item {\bf IMDB} \cite{MaasDPHNP11} is a sentiment analysis dataset and the task is to predict the sentiment (positive or negative) of a movie review.
    \item {\bf SNLI} \cite{BowmanAPM15} is a natural language inference dataset to judge the relationship between two sentences: whether the second sentence can be derived from entailment, contradiction, or neutral relationship with the first sentence.
    \item {\bf PERSPECTRUM} \cite{ChenK0CR19} is a natural language inference dataset to predict whether a relevant perspective is for/against the given claim.
        \item {\bf BoolQ} \cite{ClarkLCK0T19} is a dataset of reading comprehension instances with boolean (yes or no) answers.
    \item {\bf AG} \cite{ZhangZL15} is a sentence-level classification with regard to four news topics: World, Sports, Business, and Science/Technology.
    \item {\bf MR} \cite{PangL05} is a sentence-level sentiment classification on positive and negative movie reviews.
\end{itemize}

\subsection{Experiments on Contrastive Sets}
\label{sec:expcon}

We evaluate our model on four contrastive sets: IMDB, PERSPECTRUM, BoolQ and SNLI, which were provided by Contrast Sets\footnote{\url{https://github.com/allenai/contrast-sets}} \cite{gardner2020evaluating}. We compare our approach with BERT and RoBERTa across the original test set (Ori) and contrastive test set (Rev). \emph{Contrast consistency} (Con) is a metric defined by \citet{gardner2020evaluating} to evaluate whether a model’s predictions are all correct for the same examples in both the original test set and the contrastive test set. We fine-tune each model many times using different learning rates (1e-5,2e-5,3e-5,4e-5,5e-5) and select the best result on the contrastive test set.

From the results shown in Table \ref{tab:contrast}, we can observe that our model outperforms the baseline. Especially in the \emph{contrast consistency} metric, our method significantly outperforms other methods, which means our model is sensitive to the small change of semantic, rather than simply capturing the characteristics of the dataset. On the other hand, our model also has some improvement on the original test set, which means our method can boost the performance of PLMs on the common examples.

\begin{table}[t!]
\footnotesize
\begin{center}
\begin{tabular}{l|c|c|c|c|c}
  \toprule
  Model & Method & IMDB & AG & MR & SNLI \\
  \midrule
  \multirow{2}{*}{BERT} & Vanilla & 88.7 & 88.8 & 68.4 & 48.6 \\
  & FreeLB & 91.9 & 93.3 & 75.9 & 56.1 \\
  \midrule
  \multirow{2}{*}{RoBERTa} & Vanilla & 93.9 & 91.9 & 79.7 & 55.1 \\
  & FreeLB & 95.2 & 93.5 & 81.0 & 58.1  \\
  \midrule
  \multirow{2}{*}{CLINE} & Vanilla & \bf 94.7 & \bf 92.3 & \bf 80.4 & \bf 55.4 \\
  & FreeLB & \bf 95.9 & \bf 94.2 & \bf 82.1 & \bf 58.7 \\
  \bottomrule
\end{tabular}
\end{center}
\caption{\label{tab:adversarial} Accuracy on the adversarial test set.}
\end{table}

\subsection{Experiments on Adversarial Sets}
\label{sec:expadv}

To evaluate the robustness of the model, we compare our model with BERT and RoBERTa on the vanilla version and FreeLB version across several adversarial test sets. Instead of using an adversarial attacker to attack the model, we use the adversarial examples generated by TextFooler \cite{JinJZS20} as a benchmark to evaluate the performance against adversarial examples. TextFooler identifies the important words in the text and then prioritizes to replace them with the most semantically similar and grammatically correct words.

From the experimental results in Table \ref{tab:adversarial}, we can observe that our vanilla model achieves higher accuracy on all the four benchmark datasets compared to the vanilla BERT and RoBERTa. By constructing similar semantic adversarial examples and using the contrastive training objective, our model can concentrate the representation of the original example and the adversarial example, and then achieve better robustness. Furthermore, our method is in the pre-training stage, so it can also be combined with the existing adversarial training methods. Compared with the FreeLB version of BERT and RoBERTa, our model can achieve state-of-the-art (SOTA) performances on the adversarial sets. Experimental results on contrastive sets and adversarial sets show that our model is sensitive to semantic changes and keeps robust at the same time.

\begin{table*}[t!]
\begin{center}
\scalebox{0.9}{
\begin{tabular}{l|l|ccc|ccc|ccc}
  \toprule
  \multirow{2}{*}{Dataset} & \multirow{2}{*}{Model} & \multicolumn{3}{c}{1\%} & \multicolumn{3}{|c}{10\%} & \multicolumn{3}{|c}{100\%} \\
  & & Ori & Rev & Con & Ori & Rev & Con & Ori & Rev & Con \\
  \midrule
  \multirow{4}{*}{PERSPECTRUM} & CLINE & \bf 71.4 & \bf 60.4 & \bf 33.6 & \bf 75.1 & \bf 69.1 & \bf 55.3 & \bf 81.6 &\bf 80.2 & \bf 72.2  \\
   & w/o RTD & 67.3 & 59.4 & 29.0 & 73.4 & 67.7 & 53.0 & 81.1 & 78.3 & 68.9 \\
   & w/o Hard Negative & 59.0 & 53.0 & 14.7 & 71.4 & 68.8 & 38.2 & 80.9 & 78.2 & 65.9 \\
   & RoBERTa & 55.8 & 54.8 & 13.8 & 72.4 & 66.8 & 45.2 & 80.6 & 78.8 & 65.0 \\
   \midrule
   \multirow{4}{*}{BoolQ} & CLINE & \bf 66.7 & \bf 52.8 & \bf 33.7 & \bf 68.1 & \bf 54.0 & \bf 36.1 & \bf 73.9 & \bf 63.9 & \bf 47.8  \\
   & w/o RTD & 64.8 & 52.5 & 32.2 & 68.0 & 53.7 & 35.8 & 72.5 & 63.0 & 46.6 \\
   & w/o Hard Negative & 60.1 & 49.0 & 30.0 & 68.1 & 53.4 & 35.2 & 69.6 & 61.8 & 44.5 \\
   & RoBERTa & 60.9 & 49.3 & 27.5 & 65.2 & 53.1 & 32.8 & 69.6 & 60.6 & 43.9 \\
  \bottomrule
\end{tabular}}
\end{center}
\caption{\label{tab:ablation} Ablation study on the original test set (Ori) and contrastive test set (Rev) of PERSPECTRUM (accuracy) and BoolQ (accuracy). 1\% / 10\% indicate using 1\% / 10\% supervised training data respectively. Contrast consistency (Con) is a metric of whether a model makes correct predictions on every element in both the original test set and the contrastive test set.}
\end{table*}

\subsection{Ablation Study}
\label{ssec:as}
To further analyze the effectiveness of different factors of our CLINE, we choose PERSPECTRUM \cite{ChenK0CR19} and BoolQ \cite{ClarkLCK0T19} as benchmark datasets and report the ablation test in terms of 
1) w/o RTD: we remove the replaced token detection objective ($\mathcal{L}_{\text{RTD}}$) in our model to verify whether our model mainly benefits from the contrastive objective.
2) w/o Hard Negative: we replace the constructed negative examples with random sampling examples to verify whether the negative examples constructed by unsupervised word substitution are better. We also add 1\% and 10\% settings, meaning using only 1\% / 10\% data of the training set, to simulate a low-resource scenario and observe how the model performance across different datasets and settings. From Table \ref{tab:ablation}, we can observe that: 1) Our CLINE outperformance RoBERTa on all settings, which indicates that our method is universal and robust. Especially in the low-resource scenario (1\% and 10\% supervised training data), our method shows a prominent improvement. 2) Compared to the CLINE, w/o RTD just has a little bit of performance degradation. This proves that the improvement of performance mainly benefits from the contrastive objective and the replaced token detection objective can further make the model sensitive to the change of the words. 3) Compared to CLINE, we can see that the w/o Hard Negative has a significant performance degradation in most settings, proving the effectiveness of constructing hard negative instances.

\begin{table}[t!]
\begin{center}
\begin{tabular}{l|c|c|c}
  \toprule
  Model         & CLS  & MEAN & BS \\
  \midrule
  BERT          & 42.4 & 45.2 & 47.0 \\
  CLINE-B       & \bf 58.0 & \bf 59.2 & \bf 66.8 \\
  \midrule
  RoBERTa       & -- & 42.5 & 45.1 \\
  CLINE-R       & \bf 42.1 & \bf 42.8 & \bf 49.4 \\
  \bottomrule
\end{tabular}
\end{center}
\caption{\label{tab:hits} The max Hits(\%) on all layers of the Transformer-based encoder. We compute cosine similarity between sentence representations with the \texttt{[CLS]} token (CLS) and the mean-pooling of the sentence embedding (MEAN). And BS is short for BertScore. CLINE-B means our model trained from the BERT-base model and CLINE-R means our model trained from the RoBERTa-base model.}
\end{table}

\subsection{Sentence Semantic Representation}
\label{ssec:ssr}
To evaluate the semantic sensitivity of the models, we generate 9626 sentence triplets from a sentence-level sentiment analysis dataset MR \cite{PangL05}. Each of the triples contains an original sentence $x^{ori}$ from MR, a sentence with similar semantics $x^{syn}$ and a sentence with opposite semantic $x^{ant}$. We generate $x^{syn}$/$x^{ant}$ by replacing a word in $x^{ori}$ with its synonym/antonym from WordNet \cite{Miller95}. And then we compute the cosine similarity between sentence pairs with \texttt{[CLS]} token and the mean-pooling of all tokens. And we also use a SOTA algorithm, BertScore \cite{ZhangKWWA20} to compute similarity scores of sentence pairs. We consider cases in which the model correctly identifies the semantic relationship (e.g., if BertScore($x^{ori}$,$x^{syn}$)$
>$BertScore($x^{ori}$,$x^{ant}$)) as \emph{Hits}. And higher \emph{Hits} means the model can better distinguish the sentences, which express substantially different semantics but have few different words.

We show the max \emph{Hits} on all layers (from 1 to 12) of Transformers-based encoder in Table \ref{tab:hits}. We can observe:
1) In the BERT model, using the \texttt{[CLS]} token as sentence representation achieves worse results than mean-pooling, which shows the same conclusion as Sentence-BERT \cite{ReimersG19}. And because RoBERTa omits the NSP objective, so its result of CLS has no meaning.
2) The BertScore can compute semantic similarity better than other methods and our method CLINE-B can further improve the \emph{Hits}. 
3) By constructing positive and negative examples for contrastive learning in pre-training stage, our method CLINE-B and CLINE-R learn better sentence representation and detect small semantic changes.
4) We can observe that the RoBERTa has less \emph{Hits} than BERT, and our CLINE-B has significant improvement compared to BERT. We speculate that there may be two reasons, the first is that BERT can better identify sentence-level semantic changes because it has been trained with the next sentence prediction (NSP) objective in the pre-training stage. And the second is that the BERT is not trained enough, so it can not represent sentence semantics well, and our method can improve the semantic representation ability of the model.


\section{Related Work}
\label{sec:rw}
\subsection{Pre-trained Language Models}
\label{ssec:plm}
The PLMs have proven their advantages in capturing implicit language features. Two main research directions of PLMs are autoregressive (AR) pre-training (such as GPT \cite{radford2018improving}) and denoising autoencoding (DAE) pre-training (such as BERT \cite{DevlinCLT19}). AR pre-training aims to predict the next word based on previous tokens but lacks the modeling of the bidirectional context. And DAE pre-training aims to reconstruct the input sequences using left and right context. However, previous works mainly focus on the token-level pre-training tasks and ignore modeling the global semantic of sentences.

\subsection{Adversarial Training}
\label{ssec:at}
To make neural networks more robust to adversarial examples, many defense strategies have been proposed, and adversarial training is widely considered to be the most effective.  
Different from the image domain, it is more challenging to deal with text data due to its discrete property, which is hard to optimize. 
Previous works focus on heuristics for creating adversarial examples in the black-box setting.
\citet{BelinkovB18} manipulate every word in a sentence with synthetic or natural noise in machine translation systems.
\citet{IyyerWGZ18} leverage back-translated to produce paraphrases that have different sentence structures.
Recently, \citet{MiyatoDG17} extend adversarial and virtual adversarial training \cite{MiyatoMKI19} to text classification tasks by applying perturbations to word embeddings rather than discrete input symbols. Following this, many adversarial training methods in the text domain have been proposed and have been applied to the state-of-the-art PLMs.
\citet{li2020textat} introduce a token-level perturbation to improves the robustness of PLMs. 
\citet{ZhuCGSGL20} use the gradients obtained in adversarial training to boost the performance of PLMs. 
Although many studies seem to achieve a robust representation, our pilot experiments (Section~\ref{sec:analysis}) show that there is still a long way to go.

\subsection{Contrastive Learning}
\label{ssec:cl}
Contrastive learning is an unsupervised representation learning method, which has been widely used in learning
graph representations \cite{VelickovicFHLBH19}, 
visual representations \cite{oord2018representation,He0WXG20,ChenK0H20}, response representations \cite{LinCWLZS20,su2020dialogue}, text representations~\cite{iter2020pretraining, ding2021prototypical} and structured world models \cite{KipfPW20}.
The main idea is to learn a representation by contrasting positive pairs and negative pairs, which aims to concentrate positive samples and push apart negative samples.
In natural language processing (NLP), contrastive self-supervised learning has been widely used for learning better sentence representations.
\citet{LogeswaranL18} sample two contiguous sentences for positive pairs and the sentences from the other document as negative pairs.
\citet{luo2020capt} present contrastive pretraining for learning denoised sequence representations in a self-supervised manner. 
\citet{wu2020clear} present multiple sentence-level augmentation strategies for contrastive sentence representation learning. 
The main difference between these works is their various definitions of positive examples. 
However, recent works pay little attention to the construction of negative examples, only using simple random sampling sentences. 
In this paper, we propose a negative example construction strategy with opposite semantics to improve the sentence representation learning and the robustness of the pre-trained language model.

\section{Conclusion}
\label{sec:conclusion}
In this paper, we focus on one specific problem \textit{how to train a pre-trained language model with robustness against adversarial attacks and sensitivity to small changed semantics.} We propose CLINE, a simple and effective method to tackle the challenge. In the training phase of CLINE, it automatically generates the adversarial example and semantic negative example to the original sentence. And then the model is trained by three objectives to make full utilization of both sides of examples. Empirical results demonstrate that our method could considerably improve the sensitivity of pre-trained language models and meanwhile gain robustness.

\section*{Acknowledgments}
This research is supported by National Natural Science Foundation of China (Grant No. 61773229 and 6201101015), Tencent AI Lab Rhino-Bird Focused Research Program (No. JR202032), Shenzhen Giiso Information Technology Co. Ltd., Natural Science Foundation of Guangdong Province (Grant No. 2021A1515012640), the Basic Research Fund of Shenzhen City (Grand No. JCYJ20190813165003837), and Overseas Cooperation Research Fund of Graduate School at Shenzhen, Tsinghua University (Grant No. HW2018002).

\bibliographystyle{acl_natbib}
\bibliography{anthology,acl2021,lecbert}

\begin{thebibliography}{45}
\expandafter\ifx\csname natexlab\endcsname\relax\def\natexlab#1{#1}\fi

\bibitem[{Alzantot et~al.(2018)Alzantot, Sharma, Elgohary, Ho, Srivastava, and
  Chang}]{AlzantotSEHSC18}
Moustafa Alzantot, Yash Sharma, Ahmed Elgohary, Bo{-}Jhang Ho, Mani~B.
  Srivastava, and Kai{-}Wei Chang. 2018.
\newblock \href {https://doi.org/10.18653/v1/d18-1316} {Generating natural
  language adversarial examples}.
\newblock In \emph{Proceedings of the 2018 Conference on Empirical Methods in
  Natural Language Processing, Brussels, Belgium, October 31 - November 4,
  2018}, pages 2890--2896. Association for Computational Linguistics.

\bibitem[{Belinkov and Bisk(2018)}]{BelinkovB18}
Yonatan Belinkov and Yonatan Bisk. 2018.
\newblock \href {https://openreview.net/forum?id=BJ8vJebC-} {Synthetic and
  natural noise both break neural machine translation}.
\newblock In \emph{6th International Conference on Learning Representations,
  {ICLR} 2018, Vancouver, BC, Canada, April 30 - May 3, 2018, Conference Track
  Proceedings}. OpenReview.net.

\bibitem[{Bowman et~al.(2015)Bowman, Angeli, Potts, and Manning}]{BowmanAPM15}
Samuel~R. Bowman, Gabor Angeli, Christopher Potts, and Christopher~D. Manning.
  2015.
\newblock \href {https://doi.org/10.18653/v1/d15-1075} {A large annotated
  corpus for learning natural language inference}.
\newblock In \emph{Proceedings of the 2015 Conference on Empirical Methods in
  Natural Language Processing, {EMNLP} 2015, Lisbon, Portugal, September 17-21,
  2015}, pages 632--642. The Association for Computational Linguistics.

\bibitem[{Chen et~al.(2019)Chen, Khashabi, Yin, Callison{-}Burch, and
  Roth}]{ChenK0CR19}
Sihao Chen, Daniel Khashabi, Wenpeng Yin, Chris Callison{-}Burch, and Dan Roth.
  2019.
\newblock \href {https://doi.org/10.18653/v1/n19-1053} {Seeing things from a
  different angle: Discovering diverse perspectives about claims}.
\newblock In \emph{Proceedings of the 2019 Conference of the North American
  Chapter of the Association for Computational Linguistics: Human Language
  Technologies, {NAACL-HLT} 2019, Minneapolis, MN, USA, June 2-7, 2019, Volume
  1 (Long and Short Papers)}, pages 542--557. Association for Computational
  Linguistics.

\bibitem[{Chen et~al.(2020)Chen, Kornblith, Norouzi, and Hinton}]{ChenK0H20}
Ting Chen, Simon Kornblith, Mohammad Norouzi, and Geoffrey~E. Hinton. 2020.
\newblock \href {http://proceedings.mlr.press/v119/chen20j.html} {A simple
  framework for contrastive learning of visual representations}.
\newblock In \emph{Proceedings of the 37th International Conference on Machine
  Learning, {ICML} 2020, 13-18 July 2020, Virtual Event}, volume 119 of
  \emph{Proceedings of Machine Learning Research}, pages 1597--1607. {PMLR}.

\bibitem[{Clark et~al.(2019)Clark, Lee, Chang, Kwiatkowski, Collins, and
  Toutanova}]{ClarkLCK0T19}
Christopher Clark, Kenton Lee, Ming{-}Wei Chang, Tom Kwiatkowski, Michael
  Collins, and Kristina Toutanova. 2019.
\newblock \href {https://doi.org/10.18653/v1/n19-1300} {Boolq: Exploring the
  surprising difficulty of natural yes/no questions}.
\newblock In \emph{Proceedings of the 2019 Conference of the North American
  Chapter of the Association for Computational Linguistics: Human Language
  Technologies, {NAACL-HLT} 2019, Minneapolis, MN, USA, June 2-7, 2019, Volume
  1 (Long and Short Papers)}, pages 2924--2936. Association for Computational
  Linguistics.

\bibitem[{Devlin et~al.(2019)Devlin, Chang, Lee, and Toutanova}]{DevlinCLT19}
Jacob Devlin, Ming{-}Wei Chang, Kenton Lee, and Kristina Toutanova. 2019.
\newblock \href {https://doi.org/10.18653/v1/n19-1423} {{BERT:} pre-training of
  deep bidirectional transformers for language understanding}.
\newblock In \emph{Proceedings of the 2019 Conference of the North American
  Chapter of the Association for Computational Linguistics: Human Language
  Technologies, {NAACL-HLT} 2019, Minneapolis, MN, USA, June 2-7, 2019, Volume
  1 (Long and Short Papers)}, pages 4171--4186. Association for Computational
  Linguistics.

\bibitem[{Ding et~al.(2021)Ding, Wang, Fu, Xu, Wang, Xie, Shen, Huang, Zheng,
  and Zhang}]{ding2021prototypical}
Ning Ding, Xiaobin Wang, Yao Fu, Guangwei Xu, Rui Wang, Pengjun Xie, Ying Shen,
  Fei Huang, Hai-Tao Zheng, and Rui Zhang. 2021.
\newblock \href {https://openreview.net/forum?id=aCgLmfhIy_f} {Prototypical
  representation learning for relation extraction}.
\newblock In \emph{International Conference on Learning Representations}.

\bibitem[{Gao et~al.(2021)Gao, Yao, and Chen}]{Gao2021simcse}
Tianyu Gao, Xingcheng Yao, and Danqi Chen. 2021.
\newblock \href {http://arxiv.org/abs/2104.08821} {Simcse: Simple contrastive
  learning of sentence embeddings}.
\newblock \emph{CoRR}, abs/2104.08821.

\bibitem[{Gardner et~al.(2020)Gardner, Artzi, Basmova, Berant, Bogin, Chen,
  Dasigi, Dua, Elazar, Gottumukkala, Gupta, Hajishirzi, Ilharco, Khashabi, Lin,
  Liu, Liu, Mulcaire, Ning, Singh, Smith, Subramanian, Tsarfaty, Wallace,
  Zhang, and Zhou}]{gardner2020evaluating}
Matt Gardner, Yoav Artzi, Victoria Basmova, Jonathan Berant, Ben Bogin, Sihao
  Chen, Pradeep Dasigi, Dheeru Dua, Yanai Elazar, Ananth Gottumukkala, Nitish
  Gupta, Hannaneh Hajishirzi, Gabriel Ilharco, Daniel Khashabi, Kevin Lin,
  Jiangming Liu, Nelson~F. Liu, Phoebe Mulcaire, Qiang Ning, Sameer Singh,
  Noah~A. Smith, Sanjay Subramanian, Reut Tsarfaty, Eric Wallace, Ally Zhang,
  and Ben Zhou. 2020.
\newblock \href {https://doi.org/10.18653/v1/2020.findings-emnlp.117}
  {Evaluating models' local decision boundaries via contrast sets}.
\newblock In \emph{Proceedings of the 2020 Conference on Empirical Methods in
  Natural Language Processing: Findings, {EMNLP} 2020, Online Event, 16-20
  November 2020}, pages 1307--1323. Association for Computational Linguistics.

\bibitem[{Garg and Ramakrishnan(2020)}]{GargR20}
Siddhant Garg and Goutham Ramakrishnan. 2020.
\newblock \href {https://doi.org/10.18653/v1/2020.emnlp-main.498} {{BAE:}
  bert-based adversarial examples for text classification}.
\newblock In \emph{Proceedings of the 2020 Conference on Empirical Methods in
  Natural Language Processing, {EMNLP} 2020, Online, November 16-20, 2020},
  pages 6174--6181. Association for Computational Linguistics.

\bibitem[{He et~al.(2020)He, Fan, Wu, Xie, and Girshick}]{He0WXG20}
Kaiming He, Haoqi Fan, Yuxin Wu, Saining Xie, and Ross~B. Girshick. 2020.
\newblock \href {https://doi.org/10.1109/CVPR42600.2020.00975} {Momentum
  contrast for unsupervised visual representation learning}.
\newblock In \emph{2020 {IEEE/CVF} Conference on Computer Vision and Pattern
  Recognition, {CVPR} 2020, Seattle, WA, USA, June 13-19, 2020}, pages
  9726--9735. {IEEE}.

\bibitem[{Iter et~al.(2020)Iter, Guu, Lansing, and
  Jurafsky}]{iter2020pretraining}
Dan Iter, Kelvin Guu, Larry Lansing, and Dan Jurafsky. 2020.
\newblock \href {https://doi.org/10.18653/v1/2020.acl-main.439} {Pretraining
  with contrastive sentence objectives improves discourse performance of
  language models}.
\newblock In \emph{Proceedings of the 58th Annual Meeting of the Association
  for Computational Linguistics, {ACL} 2020, Online, July 5-10, 2020}, pages
  4859--4870. Association for Computational Linguistics.

\bibitem[{Iyyer et~al.(2018)Iyyer, Wieting, Gimpel, and
  Zettlemoyer}]{IyyerWGZ18}
Mohit Iyyer, John Wieting, Kevin Gimpel, and Luke Zettlemoyer. 2018.
\newblock \href {https://doi.org/10.18653/v1/n18-1170} {Adversarial example
  generation with syntactically controlled paraphrase networks}.
\newblock In \emph{Proceedings of the 2018 Conference of the North American
  Chapter of the Association for Computational Linguistics: Human Language
  Technologies, {NAACL-HLT} 2018, New Orleans, Louisiana, USA, June 1-6, 2018,
  Volume 1 (Long Papers)}, pages 1875--1885. Association for Computational
  Linguistics.

\bibitem[{Jiang et~al.(2020)Jiang, He, Chen, Liu, Gao, and Zhao}]{JiangHCLGZ20}
Haoming Jiang, Pengcheng He, Weizhu Chen, Xiaodong Liu, Jianfeng Gao, and Tuo
  Zhao. 2020.
\newblock \href {https://doi.org/10.18653/v1/2020.acl-main.197} {{SMART:}
  robust and efficient fine-tuning for pre-trained natural language models
  through principled regularized optimization}.
\newblock In \emph{Proceedings of the 58th Annual Meeting of the Association
  for Computational Linguistics, {ACL} 2020, Online, July 5-10, 2020}, pages
  2177--2190. Association for Computational Linguistics.

\bibitem[{Jin et~al.(2020)Jin, Jin, Zhou, and Szolovits}]{JinJZS20}
Di~Jin, Zhijing Jin, Joey~Tianyi Zhou, and Peter Szolovits. 2020.
\newblock \href {https://aaai.org/ojs/index.php/AAAI/article/view/6311} {Is
  {BERT} really robust? {A} strong baseline for natural language attack on text
  classification and entailment}.
\newblock In \emph{The Thirty-Fourth {AAAI} Conference on Artificial
  Intelligence, {AAAI} 2020, New York, NY, USA, February 7-12, 2020}, pages
  8018--8025. {AAAI} Press.

\bibitem[{Joshi et~al.(2020)Joshi, Chen, Liu, Weld, Zettlemoyer, and
  Levy}]{JoshiCLWZL20}
Mandar Joshi, Danqi Chen, Yinhan Liu, Daniel~S. Weld, Luke Zettlemoyer, and
  Omer Levy. 2020.
\newblock \href {https://transacl.org/ojs/index.php/tacl/article/view/1853}
  {Spanbert: Improving pre-training by representing and predicting spans}.
\newblock \emph{Trans. Assoc. Comput. Linguistics}, 8:64--77.

\bibitem[{Kaushik et~al.(2020)Kaushik, Hovy, and Lipton}]{KaushikHL20}
Divyansh Kaushik, Eduard~H. Hovy, and Zachary~Chase Lipton. 2020.
\newblock \href {https://openreview.net/forum?id=Sklgs0NFvr} {Learning the
  difference that makes {A} difference with counterfactually-augmented data}.
\newblock In \emph{8th International Conference on Learning Representations,
  {ICLR} 2020, Addis Ababa, Ethiopia, April 26-30, 2020}. OpenReview.net.

\bibitem[{Kipf et~al.(2020)Kipf, van~der Pol, and Welling}]{KipfPW20}
Thomas~N. Kipf, Elise van~der Pol, and Max Welling. 2020.
\newblock \href {https://openreview.net/forum?id=H1gax6VtDB} {Contrastive
  learning of structured world models}.
\newblock In \emph{8th International Conference on Learning Representations,
  {ICLR} 2020, Addis Ababa, Ethiopia, April 26-30, 2020}. OpenReview.net.

\bibitem[{Li et~al.(2020)Li, Ma, Guo, Xue, and Qiu}]{LiMGXQ20}
Linyang Li, Ruotian Ma, Qipeng Guo, Xiangyang Xue, and Xipeng Qiu. 2020.
\newblock \href {https://doi.org/10.18653/v1/2020.emnlp-main.500}
  {{BERT-ATTACK:} adversarial attack against {BERT} using {BERT}}.
\newblock In \emph{Proceedings of the 2020 Conference on Empirical Methods in
  Natural Language Processing, {EMNLP} 2020, Online, November 16-20, 2020},
  pages 6193--6202. Association for Computational Linguistics.

\bibitem[{Li and Qiu(2020)}]{li2020textat}
Linyang Li and Xipeng Qiu. 2020.
\newblock \href {http://arxiv.org/abs/2004.14543} {Textat: Adversarial training
  for natural language understanding with token-level perturbation}.
\newblock \emph{CoRR}, abs/2004.14543.

\bibitem[{Lin et~al.(2020{\natexlab{a}})Lin, Miao, Yang, Ou, Cui, Guo, and
  Miao}]{Lin0YOCGM20}
Gongqi Lin, Yuan Miao, Xiaoyong Yang, Wenwu Ou, Lizhen Cui, Wei Guo, and
  Chunyan Miao. 2020{\natexlab{a}}.
\newblock \href {https://doi.org/10.1109/ICARCV50220.2020.9305451} {Commonsense
  knowledge adversarial dataset that challenges {ELECTRA}}.
\newblock In \emph{16th International Conference on Control, Automation,
  Robotics and Vision, {ICARCV} 2020, Shenzhen, China, December 13-15, 2020},
  pages 315--320. {IEEE}.

\bibitem[{Lin et~al.(2020{\natexlab{b}})Lin, Cai, Wang, Liu, Zheng, and
  Shi}]{LinCWLZS20}
Zibo Lin, Deng Cai, Yan Wang, Xiaojiang Liu, Haitao Zheng, and Shuming Shi.
  2020{\natexlab{b}}.
\newblock \href {https://doi.org/10.18653/v1/2020.emnlp-main.741} {The world is
  not binary: Learning to rank with grayscale data for dialogue response
  selection}.
\newblock In \emph{Proceedings of the 2020 Conference on Empirical Methods in
  Natural Language Processing, {EMNLP} 2020, Online, November 16-20, 2020},
  pages 9220--9229. Association for Computational Linguistics.

\bibitem[{Liu et~al.(2019)Liu, Ott, Goyal, Du, Joshi, Chen, Levy, Lewis,
  Zettlemoyer, and Stoyanov}]{liu2019roberta}
Yinhan Liu, Myle Ott, Naman Goyal, Jingfei Du, Mandar Joshi, Danqi Chen, Omer
  Levy, Mike Lewis, Luke Zettlemoyer, and Veselin Stoyanov. 2019.
\newblock \href {http://arxiv.org/abs/1907.11692} {Roberta: {A} robustly
  optimized {BERT} pretraining approach}.
\newblock \emph{CoRR}, abs/1907.11692.

\bibitem[{Logeswaran and Lee(2018)}]{LogeswaranL18}
Lajanugen Logeswaran and Honglak Lee. 2018.
\newblock \href {https://openreview.net/forum?id=rJvJXZb0W} {An efficient
  framework for learning sentence representations}.
\newblock In \emph{6th International Conference on Learning Representations,
  {ICLR} 2018, Vancouver, BC, Canada, April 30 - May 3, 2018, Conference Track
  Proceedings}. OpenReview.net.

\bibitem[{Luo et~al.(2020)Luo, Yang, Li, Ren, and Sun}]{luo2020capt}
Fuli Luo, Pengcheng Yang, Shicheng Li, Xuancheng Ren, and Xu~Sun. 2020.
\newblock \href {http://arxiv.org/abs/2010.06351} {{CAPT:} contrastive
  pre-training for learning denoised sequence representations}.
\newblock \emph{CoRR}, abs/2010.06351.

\bibitem[{Maas et~al.(2011)Maas, Daly, Pham, Huang, Ng, and
  Potts}]{MaasDPHNP11}
Andrew~L. Maas, Raymond~E. Daly, Peter~T. Pham, Dan Huang, Andrew~Y. Ng, and
  Christopher Potts. 2011.
\newblock \href {https://www.aclweb.org/anthology/P11-1015/} {Learning word
  vectors for sentiment analysis}.
\newblock In \emph{The 49th Annual Meeting of the Association for Computational
  Linguistics: Human Language Technologies, Proceedings of the Conference,
  19-24 June, 2011, Portland, Oregon, {USA}}, pages 142--150. The Association
  for Computer Linguistics.

\bibitem[{Michel et~al.(2019)Michel, Li, Neubig, and Pino}]{MichelLNP19}
Paul Michel, Xian Li, Graham Neubig, and Juan~Miguel Pino. 2019.
\newblock \href {https://doi.org/10.18653/v1/n19-1314} {On evaluation of
  adversarial perturbations for sequence-to-sequence models}.
\newblock In \emph{Proceedings of the 2019 Conference of the North American
  Chapter of the Association for Computational Linguistics: Human Language
  Technologies, {NAACL-HLT} 2019, Minneapolis, MN, USA, June 2-7, 2019, Volume
  1 (Long and Short Papers)}, pages 3103--3114. Association for Computational
  Linguistics.

\bibitem[{Miller(1995)}]{Miller95}
George~A. Miller. 1995.
\newblock \href {https://doi.org/10.1145/219717.219748} {Wordnet: {A} lexical
  database for english}.
\newblock \emph{Commun. {ACM}}, 38(11):39--41.

\bibitem[{Miyato et~al.(2017)Miyato, Dai, and Goodfellow}]{MiyatoDG17}
Takeru Miyato, Andrew~M. Dai, and Ian~J. Goodfellow. 2017.
\newblock \href {https://openreview.net/forum?id=r1X3g2\_xl} {Adversarial
  training methods for semi-supervised text classification}.
\newblock In \emph{5th International Conference on Learning Representations,
  {ICLR} 2017, Toulon, France, April 24-26, 2017, Conference Track
  Proceedings}. OpenReview.net.

\bibitem[{Miyato et~al.(2019)Miyato, Maeda, Koyama, and Ishii}]{MiyatoMKI19}
Takeru Miyato, Shin{-}ichi Maeda, Masanori Koyama, and Shin Ishii. 2019.
\newblock \href {https://doi.org/10.1109/TPAMI.2018.2858821} {Virtual
  adversarial training: {A} regularization method for supervised and
  semi-supervised learning}.
\newblock \emph{{IEEE} Trans. Pattern Anal. Mach. Intell.}, 41(8):1979--1993.

\bibitem[{van~den Oord et~al.(2018)van~den Oord, Li, and
  Vinyals}]{oord2018representation}
A{\"{a}}ron van~den Oord, Yazhe Li, and Oriol Vinyals. 2018.
\newblock \href {http://arxiv.org/abs/1807.03748} {Representation learning with
  contrastive predictive coding}.
\newblock \emph{CoRR}, abs/1807.03748.

\bibitem[{Pang and Lee(2005)}]{PangL05}
Bo~Pang and Lillian Lee. 2005.
\newblock \href {https://doi.org/10.3115/1219840.1219855} {Seeing stars:
  Exploiting class relationships for sentiment categorization with respect to
  rating scales}.
\newblock In \emph{{ACL} 2005, 43rd Annual Meeting of the Association for
  Computational Linguistics, Proceedings of the Conference, 25-30 June 2005,
  University of Michigan, {USA}}, pages 115--124. The Association for Computer
  Linguistics.

\bibitem[{Radford et~al.(2018)Radford, Narasimhan, Salimans, and
  Sutskever}]{radford2018improving}
Alec Radford, Karthik Narasimhan, Tim Salimans, and Ilya Sutskever. 2018.
\newblock \href
  {https://www.cs.ubc.ca/~amuham01/LING530/papers/radford2018improving.pdf}
  {Improving language understanding by generative pre-training}.

\bibitem[{Reimers and Gurevych(2019)}]{ReimersG19}
Nils Reimers and Iryna Gurevych. 2019.
\newblock \href {https://doi.org/10.18653/v1/D19-1410} {Sentence-bert: Sentence
  embeddings using siamese bert-networks}.
\newblock In \emph{Proceedings of the 2019 Conference on Empirical Methods in
  Natural Language Processing and the 9th International Joint Conference on
  Natural Language Processing, {EMNLP-IJCNLP} 2019, Hong Kong, China, November
  3-7, 2019}, pages 3980--3990. Association for Computational Linguistics.

\bibitem[{Su et~al.(2020)Su, Cai, Zhou, Lin, Baker, Cao, Shi, Collier, and
  Wang}]{su2020dialogue}
Yixuan Su, Deng Cai, Qingyu Zhou, Zibo Lin, Simon Baker, Yunbo Cao, Shuming
  Shi, Nigel Collier, and Yan Wang. 2020.
\newblock \href {http://arxiv.org/abs/2012.14756} {Dialogue response selection
  with hierarchical curriculum learning}.
\newblock \emph{CoRR}, abs/2012.14756.

\bibitem[{Tan et~al.(2020)Tan, Joty, Kan, and Socher}]{TanJKS20}
Samson Tan, Shafiq~R. Joty, Min{-}Yen Kan, and Richard Socher. 2020.
\newblock \href {https://doi.org/10.18653/v1/2020.acl-main.263} {It's morphin'
  time! combating linguistic discrimination with inflectional perturbations}.
\newblock In \emph{Proceedings of the 58th Annual Meeting of the Association
  for Computational Linguistics, {ACL} 2020, Online, July 5-10, 2020}, pages
  2920--2935. Association for Computational Linguistics.

\bibitem[{Velickovic et~al.(2019)Velickovic, Fedus, Hamilton, Li{\`{o}},
  Bengio, and Hjelm}]{VelickovicFHLBH19}
Petar Velickovic, William Fedus, William~L. Hamilton, Pietro Li{\`{o}}, Yoshua
  Bengio, and R.~Devon Hjelm. 2019.
\newblock \href {https://openreview.net/forum?id=rklz9iAcKQ} {Deep graph
  infomax}.
\newblock In \emph{7th International Conference on Learning Representations,
  {ICLR} 2019, New Orleans, LA, USA, May 6-9, 2019}. OpenReview.net.

\bibitem[{Wang and Bansal(2018)}]{WangB18}
Yicheng Wang and Mohit Bansal. 2018.
\newblock \href {https://doi.org/10.18653/v1/n18-2091} {Robust machine
  comprehension models via adversarial training}.
\newblock In \emph{Proceedings of the 2018 Conference of the North American
  Chapter of the Association for Computational Linguistics: Human Language
  Technologies, NAACL-HLT, New Orleans, Louisiana, USA, June 1-6, 2018, Volume
  2 (Short Papers)}, pages 575--581. Association for Computational Linguistics.

\bibitem[{Wu et~al.(2020)Wu, Wang, Gu, Khabsa, Sun, and Ma}]{wu2020clear}
Zhuofeng Wu, Sinong Wang, Jiatao Gu, Madian Khabsa, Fei Sun, and Hao Ma. 2020.
\newblock \href {http://arxiv.org/abs/2012.15466} {{CLEAR:} contrastive
  learning for sentence representation}.
\newblock \emph{CoRR}, abs/2012.15466.

\bibitem[{Zang et~al.(2020)Zang, Qi, Yang, Liu, Zhang, Liu, and
  Sun}]{ZangQYLZLS20}
Yuan Zang, Fanchao Qi, Chenghao Yang, Zhiyuan Liu, Meng Zhang, Qun Liu, and
  Maosong Sun. 2020.
\newblock \href {https://doi.org/10.18653/v1/2020.acl-main.540} {Word-level
  textual adversarial attacking as combinatorial optimization}.
\newblock In \emph{Proceedings of the 58th Annual Meeting of the Association
  for Computational Linguistics, {ACL} 2020, Online, July 5-10, 2020}, pages
  6066--6080. Association for Computational Linguistics.

\bibitem[{Zhang et~al.(2020)Zhang, Kishore, Wu, Weinberger, and
  Artzi}]{ZhangKWWA20}
Tianyi Zhang, Varsha Kishore, Felix Wu, Kilian~Q. Weinberger, and Yoav Artzi.
  2020.
\newblock \href {https://openreview.net/forum?id=SkeHuCVFDr} {Bertscore:
  Evaluating text generation with {BERT}}.
\newblock In \emph{8th International Conference on Learning Representations,
  {ICLR} 2020, Addis Ababa, Ethiopia, April 26-30, 2020}. OpenReview.net.

\bibitem[{Zhang et~al.(2015)Zhang, Zhao, and LeCun}]{ZhangZL15}
Xiang Zhang, Junbo~Jake Zhao, and Yann LeCun. 2015.
\newblock \href
  {https://proceedings.neurips.cc/paper/2015/hash/250cf8b51c773f3f8dc8b4be867a9a02-Abstract.html}
  {Character-level convolutional networks for text classification}.
\newblock In \emph{Advances in Neural Information Processing Systems 28: Annual
  Conference on Neural Information Processing Systems 2015, December 7-12,
  2015, Montreal, Quebec, Canada}, pages 649--657.

\bibitem[{Zhu et~al.(2020)Zhu, Cheng, Gan, Sun, Goldstein, and
  Liu}]{ZhuCGSGL20}
Chen Zhu, Yu~Cheng, Zhe Gan, Siqi Sun, Tom Goldstein, and Jingjing Liu. 2020.
\newblock \href {https://openreview.net/forum?id=BygzbyHFvB} {Freelb: Enhanced
  adversarial training for natural language understanding}.
\newblock In \emph{8th International Conference on Learning Representations,
  {ICLR} 2020, Addis Ababa, Ethiopia, April 26-30, 2020}. OpenReview.net.

\bibitem[{Zhu et~al.(2015)Zhu, Kiros, Zemel, Salakhutdinov, Urtasun, Torralba,
  and Fidler}]{ZhuKZSUTF15}
Yukun Zhu, Ryan Kiros, Richard~S. Zemel, Ruslan Salakhutdinov, Raquel Urtasun,
  Antonio Torralba, and Sanja Fidler. 2015.
\newblock \href {https://doi.org/10.1109/ICCV.2015.11} {Aligning books and
  movies: Towards story-like visual explanations by watching movies and reading
  books}.
\newblock In \emph{2015 {IEEE} International Conference on Computer Vision,
  {ICCV} 2015, Santiago, Chile, December 7-13, 2015}, pages 19--27. {IEEE}
  Computer Society.

\end{thebibliography}


\end{document}